\documentclass[sigconf]{acmart}
\usepackage{todonotes}
\usepackage{footmisc}

\usepackage{amsfonts,amssymb}
\usepackage{float}
\usepackage{multirow}
\usepackage{enumitem}
\theoremstyle{definition}
\newtheorem{definition}{Definition}[section]

\AtBeginDocument{%
  }

\setcopyright{acmlicensed}
\copyrightyear{2018}
\acmYear{2018}
\acmDOI{XXXXXXX.XXXXXXX}

\acmConference[Conference acronym 'XX]{Make sure to enter the correct
  conference title from your rights confirmation emai}{June 03--05,
  2018}{Woodstock, NY}
\acmISBN{978-1-4503-XXXX-X/18/06}

\begin{document}

\title{VITA: Versatile Time Representation Learning for Temporal Hyper-Relational Knowledge Graphs}
% VITA: Versatile tIme represenTAtion

\author{ChongIn Un}
\authornote{Both authors contributed equally to this research.}

\affiliation{%
  \institution{University of Macau}
  \city{Macao}
  \country{China}
}

\author{Yuhuan Lu}
\authornotemark[1]
\affiliation{%
  \institution{University of Macau}
  \city{Macao}
  \country{China}
}

\author{Tianyue Yang}
\affiliation{%
  \institution{University of Macau}
  \city{Macao}
  \country{China}
}

\author{Dingqi Yang}
\authornote{Corresponding author.}
\affiliation{%
  \institution{University of Macau}
  \city{Macao}
  \country{China}
}

% \affiliation{%

% \affiliation{%

% \affiliation{%

% \affiliation{%

% \affiliation{%

% \affiliation{%

\renewcommand{\shortauthors}{Trovato et al.}

\begin{abstract}
Knowledge graphs (KGs) have become an effective paradigm for managing real-world facts, which are not only complex but also dynamically evolve over time. The temporal validity of facts often serves as a strong clue in downstream link prediction tasks, which predicts a missing element in a fact. Traditional link prediction techniques on temporal KGs either consider a sequence of temporal snapshots of KGs with an ad-hoc defined time interval or expand a temporal fact over its validity period under a predefined time granularity; these approaches not only suffer from the sensitivity of the selection of time interval/granularity, but also face the computational challenges when handling facts with long (even infinite) validity. Although the recent hyper-relational KGs represent the temporal validity of a fact as qualifiers describing the fact, it is still suboptimal due to its ignorance of the infinite validity of some facts and the insufficient information encoded from the qualifiers about the temporal validity. Against this background, we propose VITA, a \underline{V}ersatile t\underline{I}me represen\underline{TA}tion learning method for temporal hyper-relational knowledge graphs. We first propose a versatile time representation that can flexibly accommodate all four types of temporal validity of facts (i.e., since, until, period, time-invariant), and then design VITA to effectively learn the time information in both aspects of time value and timespan to boost the link prediction performance. We conduct a thorough evaluation of VITA compared to a sizable collection of baselines on real-world KG datasets. Results show that VITA outperforms the best-performing baselines in various link prediction tasks (predicting missing entities, relations, time, and other numeric literals) by up to 75.3\%. Ablation studies and a case study also support our key design choices.

\end{abstract}





\maketitle

% Knowledge graph and hyper-relational knowledge graph

% Temporal knowledge graph: 1) why temporal information matters, temporal validity of facts, 2) time representation of existing method, why each of them is not good. The necessity of a versatile time representation.

\section{Introduction}
Knowledge Graphs (KGs) \cite{Knowledge} have been widely recognized as a promising data management paradigm that empowers a wide range of Web applications such as question-answering \cite{KGQA}, recommendation systems \cite{KGAT}, and search engines \cite{EQFE}. A KG usually contains a large number of entities and their semantic relations under a graph structure, representing real-world facts. Traditional KGs usually represent facts as triplets \cite{TransE, TransH, TranS, TransR}, where each triplet $(s, r, o)$ connects a subject, a relation, and an object, such as \textit{(Marie Curie, Educated at, University of Paris)}. As real-world facts are often more complex and nuanced, Hyper-relational Knowledge Graphs (HKGs) \cite{HINGE, StarE, GRAN} have recently been introduced to enrich the information and enhance the semantics of a base triplet, where qualifier pairs $(k, v)$ are introduced to provide additional context to the base triplet, represented as $\{(s, r, o), \{(k_i, v_i)\}\}$. For instance, the triplet \textit{(Marie Curie, Educated at, University of Paris)} is associated with two qualifier pairs \textit{(Academic degree, Master of Science), (Academic major, Physics)} on Wikidata. These qualifiers can help distinguish the fact from similar facts by providing additional information further describing the base triplet. HKGs have demonstrated improved performance in downstream reasoning tasks \cite{HINGE, StarE, GRAN}, in particular on link prediction tasks which aim to predict a missing element (entity or relation) in a fact, such as $\{(s, r, ?), \{(k_i, v_i)\}\}$, where the question mark denotes the missing element to be predicted. Existing link prediction techniques usually resort to HKG embedding models \cite{wang2017knowledge} learning the structural information of the HKG for predicting the missing element.

%
% In real-world scenarios, many potential facts in HKGs remain unobserved due to their vast quantity. Link prediction (LP) aims to predict these missing relationships based on existing ones, thereby helping to complete the HKGs.
% \textit{\{(André Cruz, Member of sports team, Clube de Regatas do Flamengo), (number of matches played/races/starts, 26), (number of points/goals/set scored, 5)\}}
%start time 1989 and end time 1990

\begin{figure}
    \centering
    \includegraphics[width=0.95\linewidth]{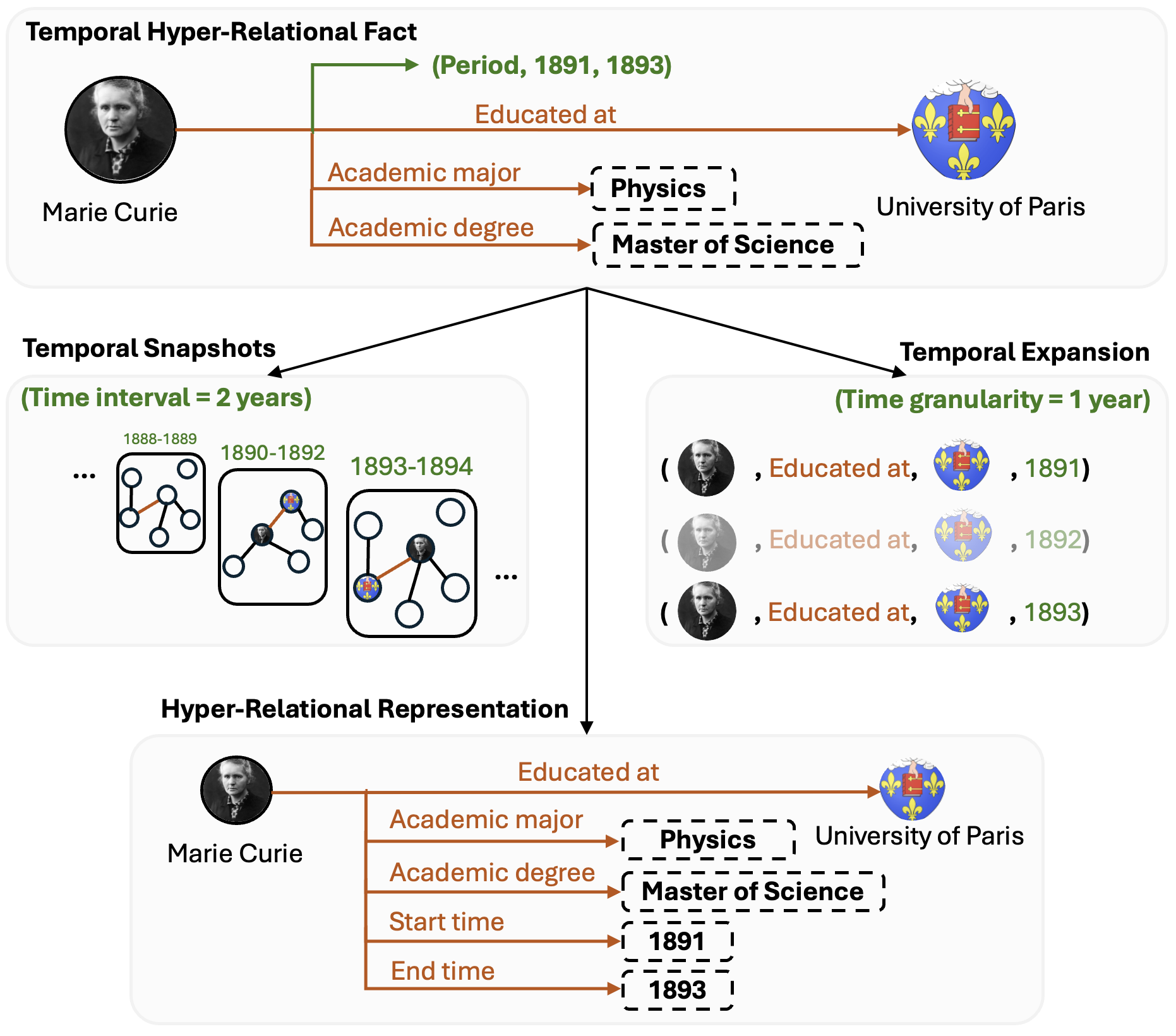}
    \vspace{-1em}
    \Description[<short description>]{<long description>}
    \caption{Different representation schemes for temporal hyper-relational facts.}
    \label{fig:IntroFact}
    \vspace{-2em}
\end{figure}

Although existing link prediction techniques on HKGs have demonstrated superiority in capturing rich semantics of the hyper-relational facts, they often overlook the temporal information of facts, which could serve as a strong clue in link prediction tasks \cite{cai2024survey, zhang2024survey}. Specifically, the temporal information of a fact usually implies the temporal validity of the fact over time. For example, as shown in Figure \ref{fig:IntroFact}, the hyper-relational fact \textit{\{(Marie Curie, Educated at, University of Paris), (Academic degree, Master of Science), (Academic major, Physics)\}} is associated with a start time 1891 and an end time 1893. When making link predictions on the subject, i.e., \textit{\{(?, Educated at, University of Paris), (Academic degree, Master of Science), (Academic major, Physics)\}}, the start and end time will serve as a strong clue for the prediction.

\label{sec:intro}

In this context, Temporal Knowledge Graph (TKG) embedding models \cite{zhang2024survey, zhang2024survey} have been widely studied to incorporate temporal information of (mostly triple) facts into the link prediction techniques. Existing works widely follow two schemes to represent and model the temporal information. First, \textit{temporal snapshots} of TKGs are created for predefined time intervals (e.g., 2 years) as shown in Figure \ref{fig:IntroFact}, where each temporal KG snapshot only contains the facts that are valid in the corresponding time interval; subsequently, the embedding models are designed to learn the semantic dynamics of the sequence of KG snapshots \cite{HGE, TeMP, T-GAP, TARGCN, DHyper}. Second, \textit{temporal expansion} duplicates a fact to multiple facts over its validity period under a predefined timestamp granularity (e.g., 1 year) as shown in Figure \ref{fig:IntroFact}; the translation-based or decomposition-based embedding models are then designed to learn the fact validity across those timestamps \cite{HyTE, BoxTE, Tero, TeAST}. However, these two schemes have their intrinsic limitations. On one hand, the ad-hoc selection of time intervals of temporal snapshots is not straightforward and yet sensitive, as the time intervals may mismatch the temporal validity of facts; a fact may be present in one or two snapshots depending on the selected time intervals, resulting in different semantic dynamics over time. On the other hand, the temporal expansion of facts not only creates redundant facts that may lead to systematic biases of oversampling from the facts of long validity, but also suffers from computational challenges in learning a large number of redundant facts, as evidenced by our experiments later. 

Recently, with the prominence of HKGs, the temporal information of facts can be naturally accommodated as the qualifiers of a fact. For example, two qualifiers, (start time, 1891) and (end time, 1893), are used to represent the validity period of a fact between the two timestamps, as shown in Figure \ref{fig:IntroFact}. Earlier HKG embedding models \cite{NaLP, RAM} usually consider such numeric-valued qualifiers the same way as other entity-valued qualifiers, and treat all of them as discrete tokens with learnable embeddings. Recent works widely recognized that these numeric-valued qualifiers should not be treated the same as other entity-valued qualifiers \cite{kristiadi2019incorporating, HyNT}, due to the intrinsic difference between nominal and numeric variables. While many subsequent studies simply remove all numeric-valued qualifiers \cite{HINGE, StarE} and focus on entity-valued qualifiers only, a few studies propose to design separate learning pipelines to better accommodate the numeric-valued qualifiers \cite{kristiadi2019incorporating, HyNT}. However, in this paper, we argue that representing the temporal information as ordinary numeric qualifiers indeed encodes insufficient information about the temporal validity of the fact. First, such representation ignores the infinite temporal validity. For example, on Wikidata, \textit{(the United States, capital, Washington, D.C.)} is associated with \textit{(start time, 1800)}, but without any end time; \textit{(the United States, part of, North America)} represents a time-invariant fact and is not associated with any time information. Second, the temporal validity of a fact implies to what extent the fact is reliable in link prediction; it is more important than other qualifiers that only provide auxiliary information about the fact, and thus requires a special design consideration in link prediction.

Against this background, we propose VITA, a \underline{V}ersatile t\underline{I}me represen\underline{TA}tion learning method for Temporal Hyper-relational Knowledge Graphs (THKGs). Specifically, we first propose a versatile time representation that can flexibly accommodate different types of temporal validity of facts. We define a time triplet consisting of a time-related conjunction $c$ followed by two time values, which can represent all four types of temporal validity: 1) valid since time $t_1$ \textit{(Since, $t_1$, $+\infty$)}, 2) valid until time $t_2$ \textit{(Until, $-\infty$, $t_2$)}, 3) valid within a period between $t_1$ and $t_2$ \textit{(Period, $t_1$, $t_2$)}, and 4) always valid or time-invariant \textit{(Invariant, $-\infty$, $+\infty$)}, where $\pm\infty$ denotes special tokens of infinity. Under this time representation scheme, a temporal hyper-relational fact is represented as $\{(s,r,o), (c, t_1, t_2), \{(k_i, v_i)\}\}$. Subsequently, VITA is designed to effectively learn from such temporal hyper-relational facts to boost link prediction performance. Following an encoder-decoder architecture, VITA designs three distinct encoders to separately encode the base triplet, the time triplet, and the set of qualifiers of a fact, where a Time Value Encoder (TVE) is designed to accommodate both time values of real-valued numbers or infinity tokens. Afterward, self-attention layers are used to capture the interaction between the three encoded representations from the three encoders, outputting the context features for the three respective decoders. Each decoder is responsible for making predictions on its corresponding elements by further integrating the context feature with the timespan information extracted from a TimeSpan Fuser (TSF), prescribing the validity timespan of facts. Finally, VITA adopts a masked training strategy and can predict any missing element in the fact with the corresponding decoder. Our contributions are summarized below:
\begin{itemize}[leftmargin=*]
    \item We revisit the existing time representation for temporal hyper-relational knowledge graphs and propose a novel versatile time representation scheme, which can flexibly accommodate different types of temporal validity of facts.
    
    % propose a new unified representation of the Temporal Hyper-relational Knowledge Graph (THKG), which includes triplets for the original relationships, qualifiers for additional semantic information, and temporal information for both time-invariant and time-sensitive facts.
    \item We proposed VITA, a \underline{V}ersatile t\underline{I}me represen\underline{TA}tion learning method for THKGs, which is designed to effectively learn from temporal hyper-relational facts under our versatile time representation to boost the link prediction performance.
    
    % to learn the representations of entities, relations and temporal information for finish the link prediction task over THKG, we also designed \todo{} focusing on capture the validity of each fact, but not only the context feature.
    \item We conduct a thorough evaluation of VITA compared to a sizable collection of baselines on real-world KG datasets. Results show that VITA outperforms the best-performing baselines in various link prediction tasks (predicting missing entities, relations, time, and numeric literals) by up to 75.3\%. Ablation studies and a case study also verify our key design choices.
    
    % The experimental results in THKG LP demonstrate that \todo{} can jointly learn from triplets and qualifiers while also capturing temporal information more effectively.
\end{itemize}

\section{Related Work}
\subsection{Embedding Models for Static KGs} 
To effectively make use of KGs, various KG embedding models have been developed to resolve link prediction tasks over static KGs, where facts are assumed to be static and do not evolve over time. Earlier studies mostly focus on predicting a missing element (entity or relation) under the triplet representation of facts $(s,r,o)$. For example, translational models use distance-based scoring functions to evaluate the plausibility of triplets \cite{TransE, TransH, TransR, TransD, TranS}; semantic matching models adopt similarity-based scoring functions to evaluate the plausibility of triplets \cite{ComplEx, DisMult, RESCAL}; deep learning models design sophisticated neural architectures to capture complex relations between entities \cite{Thomas_N._Kipf, ConvE, ConvKB, RSNs}. 

Some later works have shown that the traditional triplet representation oversimplifies the complex nature of real-world facts, as hyper-relational facts, where each fact contains multiple relations and entities, have become more and more prominent \cite{HINGE}. To learn from such hyper-relational facts, some studies adopt an n-ary representation to represent a fact as a set of relation-entity pairs ${(r_1,e_1),(r_2, e_2),...}$ and design corresponding KG embedding models to accommodate and make predictions over the facts containing an arbitrary number of entities \cite{m-TransH, RAE, GETD, HyPE, NaLP}. Moreover, some recent works have revealed that the base triplets $(s,r,o)$ indeed serve as the fundamental data structure in the KGs and preserve the essential information for link prediction, while the n-ary representation transforming the base triplets into relation-entity pairs fail to capture this key information. Subsequently, a hyper-relational representation is proposed where the base triplet can be associated with an arbitrary number of qualifiers $\{(s,r,o), \{(k_i, v_i)\}\}$, which has been widely adopted to design hyper-relational KG embedding models for link prediction tasks \cite{HINGE, StarE, GRAN, QUAD, Neuinfer, Hy-Transformer, HyNT}. 

While these models show superior performance in link prediction tasks on hyper-relational KGs, most of them focus on entities and relations only, without considering numeric literals, such as the year 1800 in the qualifier \textit{(start time, 1800)}. In this context, a few works investigate KG embedding models with numeric literals \cite{kristiadi2019incorporating, gesese2021survey, santini2022knowledge, HyNT}. While most works in this thread still follow the triplet representation\cite{kristiadi2019incorporating, gesese2021survey, santini2022knowledge}, a recent method HyNT \cite{HyNT} addresses the problem of learning from the numeric literals in the hyper-relational facts $\{(s,r,o), \{(k_i, v_i)\}\}$, where $o$ and $v$ could be numeric literals rather than entities only.

In this paper, different from these works addressing link prediction tasks over static KGs, we focus on temporal hyper-relational KGs where the temporal information of hyper-relational facts prescribes the temporal validity of the facts, and could thus serve as a strong clue for link prediction. Even though the temporal information could be treated as numeric literals and form numeric-valued qualifiers which could be handled by a few existing methods, we argue that this approach of representing time information as qualifiers indeed encodes insufficient information about the temporal validity of the facts, and thus results in suboptimal link prediction performance as evidenced by our experiments later.

\subsection{Temporal KG embedding models}
To incorporate the temporal information of facts into link prediction tasks, temporal KG embedding models have been widely investigated \cite{zhang2024survey, zhang2024survey}. Existing methods mostly assume that each fact in a KG is associated with a timestamp or a time period indicating the temporal validity of the fact. Subsequently, two representation schemes for temporal information of facts are widely adopted by existing works. On one hand, temporal snapshots of TKGs are used to represent a temporal KG as a sequence of KG snapshots, each consisting of facts that are valid under a predefined time interval (e.g., 1 year or 10 years \cite{garcia2018learning}). The embedding models are designed to learn the semantic dynamics of the sequence of KG snapshots \cite{HGE, TeMP, T-GAP, TARGCN, DHyper, li2021temporal, TiRGN, CyGNet}. For example, TeMP \cite{TeMP} learns the representations of entities and relations via message passing in a Graph Convolutional Network, with temporal dynamics being learned using Gated Recurrent Units and self-attention mechanisms; TARGCN \cite{TARGCN} learns the representations of entities and relations via message passing in a so-called temporal neighboring graph, which only allows a limited number of neighbors to pass the message and aggregates temporal dynamics via a functional time encoder \cite{xu2020inductive}. On the other hand, temporal expansion duplicates a fact to multiple facts over its validity period under a predefined timestamp granularity (e.g., 15 minutes, 24 hours, or 1 year \cite{garcia2018learning}), where each triplet is associated with one timestamp. The corresponding KG embedding models then propose various time-aware scoring functions and decomposition methods, extending traditional KG embedding models to incorporate the timestamps when making link predictions \cite{TTransE, TComplEx, DE-SimplE, ATiSE, HyTE, BoxTE, Tero, TeAST, li2021search}. For example, BoxTE \cite{BoxTE} learns the relation-specific temporal dynamics for each timestamp, and further learns the representations of entities and relations by fusing the dynamics into a scoring function \cite{abboud2020boxe}; TeRo \cite{Tero} first incorporates temporal information to the representations of entities as a rotation in the complex vector space, and then learns the representations of entities and relations together via a translation-based scoring function \cite{TransE}.

These existing works have intrinsic limitations as they not only suffer from the sensitivity of the ad-hoc selection of time interval/granularity, but also face computational challenges when handling facts with long (even infinite) validity. Moreover, as these methods all focus on the triplet representation of fact, they cannot accommodate semantic-rich qualifiers. To address these issues, we propose in this paper a versatile time presentation scheme and its corresponding embedding models for temporal hyper-relational KGs, so as to efficiently make link predictions over temporal hyper-relational facts.

\section{Versatile Time Representation} \label{sec:def}
In this section, we formally introduce our versatile time representation, followed by the corresponding temporal hyper-relational KG and link prediction tasks. First, to overcome the limitations of existing time representation schemes for temporal KGs, we propose our versatile time representation which can flexibly accommodate different types of temporal validity of facts.

\begin{definition}[\textbf{Versatile Time Triplet}]
A versatile time triplet $(c, t_1, t_2)$\footnote{We use $(c, t_1, t_2)$ to represent the time triplet without loss of generality, while $t_1$ and $t_2$ can also be infinity tokens.} defines a temporal validity of a fact, consisting of three elements: 1) a time-related conjunction $c$ specifying one of the four types of temporal validity $c \in \{Since, Until, Period, Invariant\}$, followed by two time values $t_1$ and $t_2$, where the time value can be either a real-valued number or an infinity token. Specifically, this time triplet can flexibly represent four types of time validity: 1) valid since time $t_1$ \textit{(Since, $t_1$, $+\infty$)}, 2) valid until time $t_2$ \textit{(Until, $-\infty$, $t_2$)}, 3) valid within a period between $t_1$ and $t_2$ \textit{(Period, $t_1$, $t_2$)} (or valid on a timestamp if $t_1$ equals to $t_2$), and 4) always valid or time-invariant \textit{(Invariant, $-\infty$, $+\infty$)}, where $\pm\infty$ denotes special tokens of infinity. 
\end{definition}
    
\begin{figure}
    \centering
    \includegraphics[width=\linewidth]{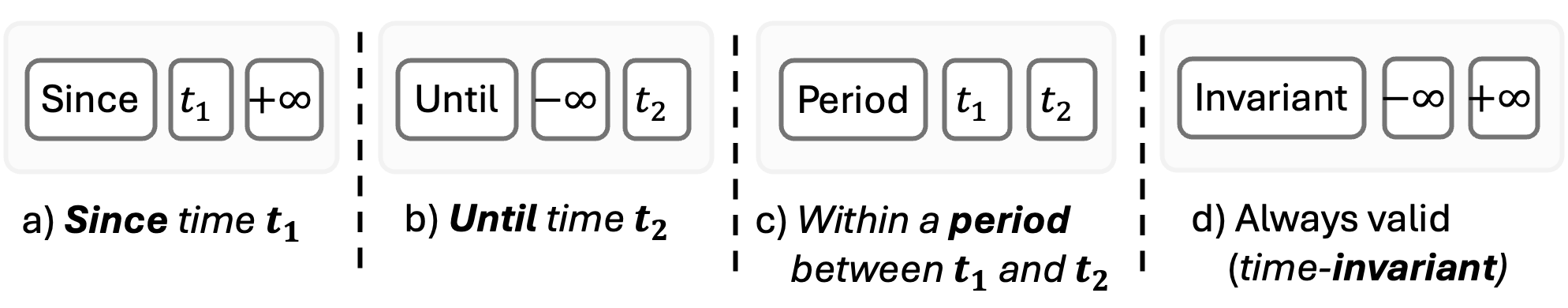}
    \vspace{-1em}
    \Description[<short description>]{<long description>}
    \caption{Four types of temporal validity represented by the versatile time triplets.}
    \label{fig:TimeConjunctionDescription}
    \vspace{-1em}
\end{figure}

\noindent Figure \ref{fig:TimeConjunctionDescription} shows the four types of temporal validity. Our versatile time triplet benefits from the following properties:
\begin{itemize}[leftmargin=*]
    \item \textbf{Preciseness.} The time values in a time triplet can be of arbitrary precision. In contrast, the existing temporal snapshots and temporal expansion schemes require a predefined time interval or a predefined timestamp granularity, respectively, which constrain the time precision.
    
    \item \textbf{Compactness.} The time triplet is a compact representation of temporal validity, compared to the existing temporal expansion scheme that results in duplicated facts suffering from both the systematic biases and computational challenges of oversampling from the facts of long validity. 
    
    \item \textbf{Completeness.} The time triplet can represent any type of temporal validity, compared to the existing HKG representation using numeric-valued qualifiers that do not account for infinite temporal validity. 
\end{itemize}
Under our versatile time representation scheme, we then define the temporal hyper-relational KG and the link prediction task below.

\begin{definition}[\textbf{Temporal Hyper-relational KG}]
A temporal hyper-relational knowledge graph (THKG) is defined as $\mathcal{G} = \{\mathcal{V}, \mathcal{R}, \mathcal{C}, \mathcal{T}, \mathcal{F}\}$, where $\mathcal{V} = \mathcal{V}_{e} \cup \mathcal{V}_n$ referring to the union of a set of entities $\mathcal{V}_{e}$ and a set of numeric literals $\mathcal{V}_n$, $\mathcal{R}$ referring to a set of relations, $\mathcal{C}$ referring to the set of time-related conjunctions, $\mathcal{T}$ referring to a set of time values, $\mathcal{F}$ referring a set of temporal hyper-relational facts. A temporal hyper-relational fact is represented as $\{(s,r,o), (c, t_1, t_2), \{(k_i, v_i)\}\}$, where $s \in \mathcal{V}_{e}$; $o, v_i \in \mathcal{V}$; $r, k_i \in \mathcal{R}$; $ c\in \mathcal{C}$; $t_1, t_2 \in \mathcal{T}$. Note that $(s,r,o)$ refers to the base triplet; $(c, t_1, t_2)$ refers to the versatile time triplet; $\{(k_i, v_i)\}$ refers to a set of qualifiers. 
\end{definition}

\begin{definition}[\textbf{Link Prediction over THKG}] \label{def:task}
The link prediction over TKHG is to predict a missing element in a temporal hyper-relational fact $\{(s,r,o), (c, t_1, t_2), \{(k_i, v_i)\}\}$. The missing element could be any element $s, r, o, c, t_1, t_2, k_i$, or $v_i$ in this fact. Depending on the position and modality of the missing element, link prediction tasks can be classified into 1) entity prediction where the missing element is one of $s, o, v_i \in \mathcal{V}_{e}$; 2) relation prediction where the missing element is one of $r$, $k_i$ or $c$; 3) time prediction where the missing element is one of $t_1$ or $t_2$; 4) numeric literal prediction where the missing element is one of $o, v_i \in \mathcal{V}_{n}$. When predicting time values, there are two primary types of reasoning \cite{chen2024unified}: interpolation, which involves predicting missing links at any historical timestamp, and extrapolation, which involves predicting future links based on historical timestamps. This paper focuses on the interpolation setting.
\end{definition}

\begin{figure}[t]
    \centering
    \includegraphics[width=\linewidth]{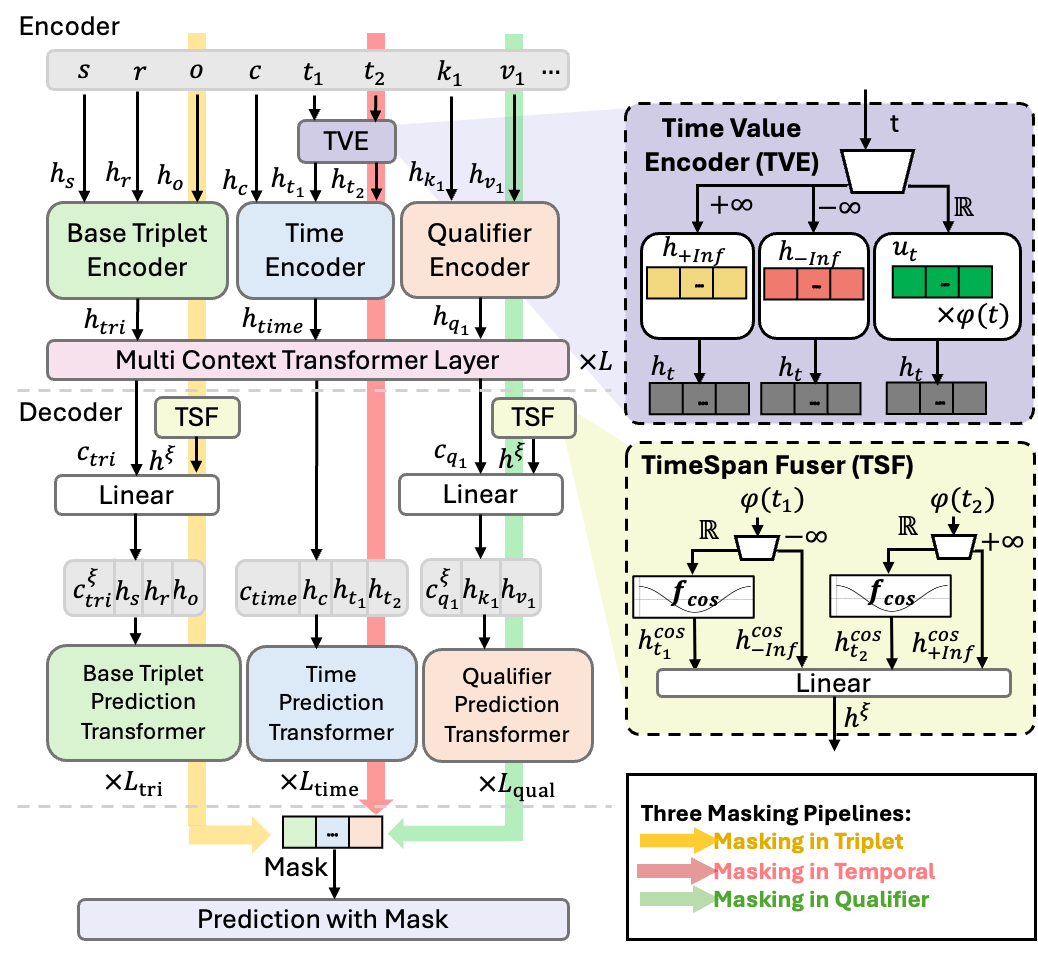}
    \vspace{-1.5em}
    \Description[<short description>]{<long description>}
    \caption{Overview of VITA.}
    \label{fig:Model}
    \vspace{-1.5em}
\end{figure}

\section{VITA}
The proposed VITA model consists of three key modules, designed to directly learn from temporal hyper-relational facts with our versatile time representation. First, it utilizes three distinct encoders to separately encode the base triplet, the time triplet, and the set of qualifiers of a fact, where a Time Value Encoder (TVE) is proposed to accommodate both time values of real-valued numbers or infinity tokens. Self-attention layers are then employed to capture the correlations among the three encoded representations, generating three context features. Second, it exploits three decoders integrating the respective context features with the timespan information extracted from a TimeSpan Fuser (TSF), prescribing the validity timespan of facts. Finally, it makes predictions with masks for the link prediction tasks. The architecture of VITA is shown in Figure \ref{fig:Model} and we present details of the three modules below.

\subsection{Encoder}
\label{sec:Encoder}
In a temporal hyper-relational fact $\{(s,r,o), (c, t_1, t_2), \{(k_i, v_i)\}\}$, the base triplet, the time triplet, and the set of qualifiers support the fact from different perspectives \cite{HINGE}. To this end, we separately model the three components with a base triplet encoder, a time encoder, and a qualifier encoder, respectively, as shown in Figure \ref{fig:Model}. Specifically, we employ three distinct linear layers to encode the structural information within each component, thereby generating three respective representations:

% TODO: add lu2023schema in camera-ready version

% \todo{} learns the representations of each entity, relation and temporal information first, and further learns the context information between different part of THKG fact. Given a temporal hyper-relational fact $((s,r,o), (c,t_1,t_2), \{(k_i,v_i)\}_{i=1}^{n})$, where $n$ refer to the numbers of qualifier pairs, from THKG $\mathcal{G}$. \todo{} The encoded structural vector $\mathbf{h_{pri}}$ for primary triplet, $\mathbf{h_{time}}$ for temporal information, $\mathbf{h_{q_i}}$ for the i-th pair of qualifier are computed.

\begin{equation}
    \begin{split}
        \mathbf{h}_{tri} &= \mathbf{W}_{tri}(\mathbf{h}_s\parallel\mathbf{h}_r\parallel\mathbf{h}_o)+\mathbf{b}_{tri} \\
        \mathbf{h}_{time} &= \mathbf{W}_{time}(\mathbf{h}_c\parallel\mathbf{h}_{t_1}\parallel\mathbf{h}_{t_2})\ +\mathbf{b}_{time} \\
        \mathbf{h}_{q_i} &= \mathbf{W}_{qual}(\mathbf{h}_{k_i}\parallel\mathbf{h}_{v_i})+\mathbf{b}_{qual}
    \end{split}
\end{equation}

\noindent where $\parallel$ denotes a concatenation operation. We present the details for each encoded element below.

First, the \textit{time encoder} for the time triplet should be designed to accommodate both time values of real-valued numbers or infinity tokens. To this end, we first design a \textbf{Time Value Encoder (TVE)} as shown in Figure \ref{fig:Model}. Specifically, for real-valued $t$, $\mathbf{h}_{t_1} = \mathbf{u}_t \times \varphi(t_1)$ and $\mathbf{h}_{t_2} = \mathbf{u}_t \times \varphi(t_2)$ denote the embedding vectors of $t_1$ and $t_2$ respectively, wherein $\mathbf{u}_t \in \mathbb{R}^d$ is a time unit embedding vector and $\varphi(\cdot)$ is a min-max normalization function for all real-valued time values. This design choice integrates the linearity of numeric literals into learnable embeddings. In the case of infinity tokens $\pm\infty$, we directly adopt two embedding vectors $\mathbf{h}_{-\infty}, \mathbf{h}_{+\infty} \in \mathbb{R}^d$ for the corresponding $\mathbf{h}_{t_1}$ or $\mathbf{h}_{t_2}$. Afterward, $\mathbf{h}_{c} \in \mathbb{R}^d$ denotes the embedding vector of time-related conjunction $c$, which are fed together with $\mathbf{h}_{t_1}$ and $\mathbf{h}_{t_2}$ to a linear layer. Subsequently, under this design, the encoder can flexibly accommodate different types of temporal validity of facts.

Second, the \textit{base triplet encoder} and \textit{qualifier encoder} are designed in a similar way. Specifically, $\mathbf{h}_s, \mathbf{h}_o$ and $\mathbf{h}_{v_i}\in \mathbb{R}^d$ denote the embedding vectors of entity $s, o$ and $v_i \in \mathcal{V}_e$ respectively; in case of numeric literals $o, v_i \in \mathcal{V}_n$, we compute $\mathbf{h}_{o} = \mathbf{u}_r \times \varphi_r(o)$ and $\mathbf{h}_{v_i} = \mathbf{u}_{k_i} \times \varphi_{k_i}(v_i)$ where $\mathbf{u}_r, \mathbf{u}_{k_i} \in \mathbb{R}^d$ is relation unit embedding vectors and $\varphi_r(\cdot)$ is a relation-specific min-max normalization function.  $\mathbf{h}_r, \mathbf{h}_{k_i}\in \mathbb{R}^d$ denote the embedding vector of relation $r$ and $k_i$ respectively. Subsequently, $\mathbf{h}_s, \mathbf{h}_r, \mathbf{h}_o$ are fed to the base triplet encoder. $\mathbf{h}_{k_i}$ and $\mathbf{h}_{v_i}$ are fed to the qualifier encoder. Note that the qualifier encoder is applied to each qualifier pair of a temporal hyper-relational fact.

% If given $o \in \mathcal{V}_n$ or $v_i \in \mathcal{V}_n$, we will compute $\mathbf{h}_{o} = \mathbf{u}_r \times \varphi_r(o)$ and $\mathbf{h}_{v_i} = \mathbf{u}_{k_i} \times \varphi_{k_i}(v_i)$ where $\mathbf{u}_r \in \mathbb{R}^d$ and $\mathbf{u}_{v_i} \in \mathbb{R}^d$ is a relation unit embedding vector. 

% Notice that, we will replace $\mathbf{h}_{t_1}$ to $\mathbf{h}_{-\infty} \in \mathbb{R}^d$ if $t_1$ is infinity, $\mathbf{h}_{t_2}$ to $\mathbf{h}_{+\infty}\in \mathbb{R}^d$ if $t_2$ is infinity. 

Finally, we use self-attention layers to capture the interactions between the three encoded representations \cite{transformer}, generating three context features for the base triplet, the time triplet and the qualifiers, denoted as $\mathbf{c}_{tri}, \mathbf{c}_{time}, \mathbf{c}_{q_i}$, respectively.

\subsection{Decoder}
Upon obtaining the three context features, we develop three respective decoders, which only focus on making predictions on the corresponding elements within the base triplet, the time triplet, and the qualifiers, respectively.

First, to make predictions on the base triplet and the qualifiers, the fact validity timespan serves as a strong clue for identifying plausible answers. In this context, we design a \textbf{TimeSpan Fuser (TSF)} to further integrate the validity timespan information into the context features, as shown in Figure \ref{fig:Model}. To this end, we leverage the translation-invariant time encoding technique proposed by \cite{xu2020inductive}, which is designed to represent the real-valued relative timespan via a series of learnable sinusoidal functions; and we extend it to accommodate also the infinity tokens in our case. Specifically, if $t_1$ or $t_2$ are not infinity, we compute the features $h^{cos}_{t_1}$ and $h^{cos}_{t_2}$ as:
\begin{equation}
        \mathbf{h}^{cos}_{t_1} = \frac{1}{\sqrt{d}}\mathbf{cos}(\varphi(t_1)\times \mathbf{\omega} + \mathbf{\phi}), 
        \mathbf{h}^{cos}_{t_2} = \frac{1}{\sqrt{d}}\mathbf{cos}(\varphi(t_2)\times \mathbf{\omega} + \mathbf{\phi})
\end{equation}
where $\mathbf{cos}(\cdot)$ denote the cosine function, $\omega\in\mathcal{R}^d$ and $\phi \in \mathcal{R}^d$ denote the learnable angular frequency and phase constant respectively. In the case of $t_1$ or $t_2$ being an infinity token $\pm\infty$, we assign two embedding vectors $\mathbf{h}^{cos}_{-\infty}$ and $\mathbf{h}^{cos}_{+\infty}$ for the corresponding infinity token. Afterward, we adopt a linear layer to learn the timespan specified by $\mathbf{h}^{cos}_{t_1}$ and $\mathbf{h}^{cos}_{t_2}$, outputting the timespan feature $\mathbf{h}^{\xi}$. Finally, we integrate the timespan feature $\mathbf{h}^{\xi}$ into the context features $\mathbf{c}_{tri}$ and $\mathbf{c}_{q_i}$, outputting the timespan fused context features $\mathbf{c}^{\xi}_{tri}$ and $\mathbf{c}^{\xi}_{q_i}$ for the base triplet and qualifiers, respectively. Note that this timespan encoder is not applied to the context feature for the time triplet $\mathbf{c}_{time}$ to avoid information leakage when making predictions on elements of the time triplet.

% To enhance the exact representation of the triplet and its affiliated qualifiers with the temporal validity of the fact, we resort to the cosine function to encode the time period \cite{ding2023exploring}. 

% consider that an entire time series can be projected into a cosine curve, so that different time periods can be represented as a vector universally by employing a cosine function. Then, the validity of contextualized vector of each fact can be learnt by fusing the temporal feature together. \todo{} 

% \begin{equation}
%     \begin{split}
%         \mathbf{h}^{cos}_{t_1} &= \frac{1}{\sqrt{d}}\mathbf{cos}(\varphi(t_1)\times \mathbf{\omega} + \mathbf{\phi})\\
%         \mathbf{h}^{cos}_{t_2} &= \frac{1}{\sqrt{d}}\mathbf{cos}(\varphi(t_2)\times \mathbf{\omega} + \mathbf{\phi})\\
%     \end{split}
% \end{equation}

To make predictions on the elements at different positions of a temporal hyper-relational fact, we design three different pipelines:
\begin{itemize}[leftmargin=*]
    \item For the base triplet, we feed the timespan fused context feature $\mathbf{c}^{\xi}_{tri}$ with $\mathbf{h}_s, \mathbf{h}_r$, and $\mathbf{h}_o$ to a \textit{base triplet prediction transformer} and further obtain the encoded representation $\mathbf{y}_{s}, \mathbf{y}_{r}$, and $\mathbf{y}_{o}$, ready for predicting the three respective elements $s, r$ and $o$.

    \item For the qualifiers, we adopt a similar process as the base triplet, feeding the timespan fused context feature $\mathbf{c}^{\xi}_{q_i}$ with $\mathbf{h}_{k_i}$ and $\mathbf{h}_{v_i}$ into a \textit{qualifier prediction transformer} and outputting $\mathbf{y}_{k_i}$ and $\mathbf{y}_{v_i}$, ready for predicting the two respective elements $k_i$ and $v_i$. 

    \item For the time triplet, we directly feed the context feature $\mathbf{c}_{time}$ (without TSE) with $\mathbf{h}_c, \mathbf{h}_{t_1}$, and $\mathbf{h}_{t_2}$ to a \textit{time prediction transformer}, outputting the encoded representation $\mathbf{y}_{c}, \mathbf{y}_{t_1}$, and $\mathbf{y}_{t_2}$, ready for predicting the three respective elements $c, t_1$ and $t_2$.
\end{itemize}

\subsection{Prediction with Mask}
We adopt a masked training strategy \cite{devlin2018bert} for training VITA. Specifically, given a THKG query fact with a missing element to be predicted, we first use a mask embedding vector $\mathbf{x}_{mask}$ to represent the missing element, then feed the masked query fact into VITA, and finally get the corresponding output $\mathbf{y}_{mask}$. Depending on the position and modality of the missing element in a query fact, the masked element could be an entity $s, o, v_i \in \mathcal{V}_{e}$, a numeric literal $o, v_i \in \mathcal{V}_{n}$, a relation $r, k_i \in \mathcal{R}$, a time-related conjunction $c \in \mathcal{C}$, or a time value $t_1, t_2 \in \mathcal{T}$. Subsequently, we adopt individual prediction layers for each case above.

On the one hand, to predict tokens (including entities, relations, and time-relation conjunctions), we adopt a linear layer with softmax activation to output the probability of the potential elements being the correct answer. Taking the entity prediction as an example, after obtaining the output embedding vector $\mathbf{y}_{mask}$, we compute the probability $\mathbf{P}_{ent}$ over all entities as:
\begin{equation}
\label{eq:mask}
    \mathbf{P}_{ent} = \mathbf{softmax}(\mathbf{y}_{mask}\mathbf{W}_{ent} + \mathbf{b}_{ent})
\end{equation}
where $\mathbf{W}_{ent} \in \mathbb{R}^{d \times |\mathcal{V}_{e}|}, \mathbf{b}_{ent} \in \mathbb{R}^{|\mathcal{V}_{e}|}$ denote the learnable weight and bias of the linear layer. Subsequently, we pick the entity with the highest probability as the predicted entity. Similar prediction processes are applied to relations and time-related conjunctions.

% Correspondingly, we compute the cross-entropy loss against the ground truth entity, denoted as $\mathcal{L}_{ent}$. Similar learning processes are applied to relations and time-related conjunctions, resulting in two respective losses $\mathcal{L}_{rel}$ and $\mathcal{L}_{conj}$.

On the other hand, to predict numeric values (including time values\footnote{We do not predict the infinity token, because the time-relation conjunction uniquely determines the infinity token in a fact.} and numeric literals), we adopt a linear layer without activation to output a real-valued number as the predicted value. Taking time prediction as an example, after obtaining the output embedding vector $\mathbf{y}_{mask}$, we compute the predicted time $\mathbf{P}_{time}$ as:
\begin{equation}
\label{eq:time}
    \mathbf{P}_{time} = \mathbf{y}_{mask}\mathbf{W}_{time} + \mathbf{b}_{time}
\end{equation}
where $\mathbf{W}_{time} \in \mathbb{R}^{d \times 1}, \mathbf{b}_{time} \in \mathbb{R}$ denote the learnable weight and bias. A similar learning process is applied to numeric literals.

\begin{table*}
\caption{Dataset statistics. $N_{train}$, $N_{valid}$ and $N_{test}$ denote the number of facts in training, valid and test sets respectively; $|\mathcal{V}|$ denote the total number of entities; $|\mathcal{R}|$ denote the number of relations; $T_{since}\%$, $T_{until}\%$, $T_{period}\%$ and $T_{invariant}\%$ denote the percentage of fact having each of the four types of temporal validity, i.e., \textit{Since}, \textit{Until}, \textit{Period}, and \textit{Invariant}) respectively; $qual\%$ denotes the percentage of facts containing at least one qualifier.}
\vspace{-1em}
\label{tbl:DatasetStatistic}
% \begin{tabular}{c|cccccccccc}
\small
\begin{tabular}{c|rrrrrrrrrr}

\hline
Dataset    & $N_{train}$ & $N_{valid}$ & $N_{test}$ & $|\mathcal{V}|$ & $|\mathcal{R}|$ & $T_{since}\%$ & $T_{until}\%$ & $T_{period}\%$ & $T_{invariant}\%$ & $qual\%$  \\ \hline
Wiki       & 53,088    & 6,636     & 6,639    & 13,645    & 168      & 3.99\%    & 1.54\% & 94.47\% & 0        & 63.23\% \\
YAGO       & 16,112    & 2,014     & 2,022    & 10,959    & 48       & 5.09\%    & 0.72\% & 94.19\% & 0        & 24.26\% \\
wikipeople & 43,068    & 5,215     & 5,213    & 21,183    & 174      & 12.18\%   & 2.56\% & 82.89\% & 2.37\%   & 35.84\% \\
ICEWS14 & 72,584 & 9,073  & 9,073 & 7,128 & 230 & 0	& 0 & 100.00\% & 0 & 0 \\ \hline
\end{tabular}
\end{table*}

\subsection{Model Training}
Our model is trained to be able to make predictions on any missing element in a query fact. To this end, we iterate over all elements in a training fact, and treat each element as a masked position to make predictions. If the masked position is a token (i.e., entities, relations, and time-relation conjunctions), we compute the cross-entropy loss against the ground truth token, resulting in three respective losses, denoted as $\mathcal{L}_{ent}$, $\mathcal{L}_{rel}$, and $\mathcal{L}_{conj}$. If the masked position is a numeric value (i.e., time values and numeric literals), we compute the Mean Squared Error (MSE) loss against the ground truth value, resulting in two respective losses $\mathcal{L}_{num}$ and $\mathcal{L}_{time}$. Finally, the overall loss $\mathcal{L}$ of VITA is the sum of these losses:
\begin{equation}
    \mathcal{L} = \mathcal{L}_{ent} + \mathcal{L}_{rel} + \mathcal{L}_{conj} + \lambda(\mathcal{L}_{num} + \mathcal{L}_{time})
\end{equation}

% \begin{equation}
%     \mathcal{L} = \lambda_{ent}\mathcal{L}_{ent} + \lambda_{rel}\mathcal{L}_{rel} + \lambda_{conj}\mathcal{L}_{conj} + \lambda_{num}\mathcal{L}_{num} + \lambda_{time}\mathcal{L}_{time}
% \end{equation}

\noindent where $\lambda$ is used to balance the cross-entropy losses and the MSE losses. However, we find in our experiments that the overall best settings are $\lambda=1$ on most datasets, which also aligns with the findings in \cite{HyNT}; probably because the distribution of missing elements over the five types is consistent in training/valid/test datasets.

\begin{table*}[t]
\caption{Link prediction performance on entities. Note that ``out of memory'' errors are observed in a few cases on our benchmarking hardware (Intel Xeon5320@2.20GHz, 256GB RAM@3200Hz, NVIDIA GeForce RTX 4090 24GB, Ubuntu 18.04).}
\vspace{-1em}
\label{tab:AllLinkPrediction}
% \small
% \footnotesize
\resizebox{\linewidth}{!}{
\begin{tabular}{ll|cccc|cccc|cccc|cccc}
\hline
\multicolumn{2}{c|}{\multirow{2}{*}{Method}} & \multicolumn{4}{c|}{Wiki} & \multicolumn{4}{c|}{YAGO} & \multicolumn{4}{c|}{wikipeople} & \multicolumn{4}{c}{ICEWS14} \\ \cline{3-18} 
\multicolumn{2}{c|}{} & MRR & Hit@1 & Hit@3 & Hit@10 & MRR & Hit@1 & Hit@3 & Hit@10 & MRR & Hit@1 & Hit@3 & Hit@10 & MRR & Hit@1 & Hit@3 & Hit@10 \\ \hline
\multicolumn{1}{l|}{\multirow{9}{*}{\begin{tabular}[c]{@{}c@{}}TKG\\ Family\end{tabular}}} & TeRo & 0.1516 & 0.0782 & 0.1576 & 0.3005 & 0.0743 & 0.0326 & 0.0845 & 0.1680 & 0.1492 & 0.0644 & 0.1503 & 0.3164 & 0.4564 & 0.3837 & 0.4895 & 0.6382 \\
\multicolumn{1}{l|}{} & DE-SimplE & 0.1510 & 0.0637 & 0.1624 & 0.3288 & 0.1648 & 0.1037 & 0.1651 & 0.2834 & \multicolumn{4}{c|}{Out of memory} & 0.4247 & 0.3658 & 0.4696 & 0.6092 \\
\multicolumn{1}{l|}{} & BoxTE & 0.1736 & 0.0983 & 0.1894 & 0.3590 & 0.1439 & 0.1029 & 0.1637 & 0.2684 & 0.1537 & 0.0588 & 0.1565 & 0.3764 & 0.4895 & 0.4237 & 0.5254 & 0.6569 \\
\multicolumn{1}{l|}{} & HypeTKG &  \multicolumn{4}{c|}{Out of memory} & 0.0872 & 0.0549 & 0.0901 & 0.1426 &  \multicolumn{4}{c|}{Out of memory} & 0.4983 & \underline{0.4275} & 0.5326 & 0.6359  \\ \cline{2-18}
\multicolumn{1}{l|}{} & HGE & 0.1128 & 0.0625 & 0.1317 & 0.2649 & 0.1327 & 0.0982 & 0.1446 & 0.2583 & 0.2085 & 0.1281 & 0.2066 & 0.4031 & \underline{0.5166} & 0.4218 & \underline{0.5815} & 0.6743 \\
\multicolumn{1}{l|}{} & TARGCN & 0.1520 & 0.0860 & 0.1628 & 0.2762 & 0.1369 & 0.1013 & 0.1375 & 0.2002 & 0.1994 & 0.1306 & 0.2458 & 0.3517 & 0.4886 & 0.3832 & 0.5500 & 0.6841 \\ \cline{2-18}
\multicolumn{1}{l|}{} & literalE-DisMult & 0.0968 & 0.0489 & 0.1046 & 0.1973 & 0.0908 & 0.0607 & 0.0896 & 0.1452 & 0.0804 & 0.0452 & 0.0808 & 0.1472 & 0.0473 & 0.0180 & 0.0395 & 0.0959 \\
\multicolumn{1}{l|}{} & literalE-ComplEx & 0.0892 & 0.0466 & 0.095 & 0.1806 & 0.0826 & 0.0580 & 0.0830 & 0.1314 & 0.0760 & 0.0410 & 0.0781 & 0.1462 & 0.0369 & 0.0129 & 0.0324 & 0.0797 \\
\multicolumn{1}{l|}{} & literalE-ConvE & 0.0796 & 0.0468 & 0.0833 & 0.1447 & 0.0741 & 0.0525 & 0.0734 & 0.1170 & 0.0739 & 0.0427 & 0.0763 & 0.1357 & 0.0375 & 0.0129 & 0.0316 & 0.0781 \\ 	
\hline
\multicolumn{1}{l|}{\multirow{9}{*}{\begin{tabular}[c]{@{}c@{}}HKG\\ Family\end{tabular}}} & NaLP-Fix & 0.0450 & 0.0213 & 0.0424 & 0.0823 & 0.0955 & 0.0778 & 0.0999 & 0.1262 & 0.0922 & 0.0604 & 0.0957 & 0.1508 & 0.2114 & 0.1324 & 0.2314 & 0.3700 \\
\multicolumn{1}{l|}{} & HINGE & 0.1081 & 0.0601 & 0.1097 & 0.1943 & 0.0860 & 0.0508 & 0.0880 & 0.1485 & 0.1374 & 0.0827 & 0.1432 & 0.2462 & 0.2508 & 0.1554 & 0.2801 & 0.4476 \\
\multicolumn{1}{l|}{} & StarE & 0.1387 & 0.0732 & 0.1418 & 0.2721 & 0.1105 & 0.0657 & 0.1202 & 0.2000 & 0.1945 & 0.1318 & 0.2095 & 0.3208 & 0.2386 & 0.1611 & 0.2609 & 0.3964 \\
\multicolumn{1}{l|}{} & GRAN & 0.2796 & 0.2204 & 0.2992 & 0.3932 & 0.1675 & 0.1163 & 0.1790 & 0.2715 & \underline{0.2392} & \underline{0.1723} & \underline{0.2633} & 0.3731 & 0.1894 & 0.1148 & 0.2101 & 0.3445 \\
\multicolumn{1}{l|}{} & HyConvE & 0.2768 & 0.2038 & 0.2884 & 0.4233 & \underline{0.1815} & \underline{0.1241} & 0.1905 & \textbf{0.3055} & 0.1868 & 0.1200 & 0.1982 & 0.3216 & 0.2194 & 0.1203 & 0.2389 & 0.4347 \\
\multicolumn{1}{l|}{} & HypE & 0.2279 & 0.1625 & 0.2420 & 0.3534 & 0.1549 & 0.1038 & 0.1698 & 0.2546 & 0.1496 & 0.0903 & 0.1621 & 0.2686 & 0.2645 & 0.1670 & 0.2916 & 0.4707 \\
\multicolumn{1}{l|}{} & HyNT & \underline{0.3609} & \underline{0.2811} & \underline{0.3795} & \textbf{0.5244} & 0.1792 & 0.1286 & \underline{0.1930} & 0.2812 & 0.2297 & 0.1574 & 0.2452 & \underline{0.3839} & 0.4728 & 0.3600 & 0.5336 & \underline{0.6908} \\
% \multicolumn{1}{l|}{} & HAHE & 0.3018 & 0.2364 & 0.3211 & 0.4302 & 0.2361 & 0.1844 & 0.2546 & 0.3404 & 0.2881 & 0.2152 & 0.3152 & 0.4326 & 0.4563 & 0.3722 & 0.5664 & 0.6431 \\
\hline
\multicolumn{1}{l|}{Ours} & VITA & \textbf{0.3780} & \textbf{0.3113} & \textbf{0.3926} & \underline{0.5168} & \textbf{0.1898} & \textbf{0.1366} & \textbf{0.2049} & \underline{0.2993} & \textbf{0.2631} & \textbf{0.2012} & \textbf{0.2801} & \textbf{0.3897} & \textbf{0.5450} & \textbf{0.4396} & \textbf{0.6132} & \textbf{0.7361} \\ \hline
\end{tabular}
}
\end{table*}

\begin{table*}[t]
\caption{Link prediction performance on relations. We exclude the methods that cannot predict relations.}
\vspace{-1em}
\label{tab:AllRelationPreiction}
% \small
\resizebox{\linewidth}{!}{
\begin{tabular}{ll|cccc|cccc|cccc|cccc}
\hline
\multicolumn{2}{c|}{\multirow{2}{*}{Method}} & \multicolumn{4}{c|}{Wiki} & \multicolumn{4}{c|}{YAGO} & \multicolumn{4}{c|}{wikipeople} & \multicolumn{4}{c}{ICEWS14} \\ \cline{3-18} 
\multicolumn{2}{c|}{} & \multicolumn{1}{l}{MRR} & Hit@1 & Hit@3 & Hit@10 & \multicolumn{1}{l}{MRR} & Hit@1 & Hit@3 & Hit@10 & \multicolumn{1}{l}{MRR} & Hit@1 & Hit@3 & Hit@10 & \multicolumn{1}{l}{MRR} & Hit@1 & Hit@3 & Hit@10 \\ \hline
\multicolumn{1}{c|}{\multirow{5}{*}{\begin{tabular}[c]{@{}c@{}}HKG\\ Family\end{tabular}}} & NaLP-Fix & \multicolumn{1}{l}{0.1509} & 0.0791 & 0.1651 & 0.2857 & \multicolumn{1}{l}{0.2799} & 0.1435 & 0.2722 & 0.4955 & \multicolumn{1}{l}{0.7701} & 0.6777 & 0.8356 & 0.9400 & 0.0185 & 0.0013 & 0.0065 & 0.0303 \\
\multicolumn{1}{c|}{} & HINGE & \multicolumn{1}{l}{0.9661} & 0.9531 & 0.9761 & 0.9887 & \multicolumn{1}{l}{0.8826} & 0.8017 & \underline{0.9613} & \textbf{0.9936} & \multicolumn{1}{l}{0.9262} & 0.8963 & 0.9485 & 0.9711 & 0.3449 & 0.2024 & 0.3889 & 0.6696 \\
\multicolumn{1}{c|}{} & GRAN & \textbf{0.9861} & \underline{0.9802} & \textbf{0.9913} & \textbf{0.9948} & \multicolumn{1}{l}{0.9119} & 0.8678 & 0.9492 & 0.9813 & \underline{0.9530} & \underline{0.9359} & \underline{0.9659} & \underline{0.9781} & 0.3570 & 0.2288 & 0.4182 & 0.6118 \\
\multicolumn{1}{c|}{} & HyNT & \multicolumn{1}{l}{0.9800} & 0.9725 & 0.9879 & \underline{0.9903} & \textbf{0.9483} & \underline{0.9214} & \textbf{0.9720} & \underline{0.9850} & \multicolumn{1}{l}{0.9436} & 0.9268 & 0.9582 & 0.9657 & \textbf{0.7844} & \underline{0.7196} & \textbf{0.8279} & \textbf{0.9052} \\
\hline
\multicolumn{1}{l|}{Ours} & VITA & \underline{0.9856} & \textbf{0.9823} & \underline{0.9880} & 0.9897 & \underline{0.9462} & \textbf{0.9346} & 0.9531 & 0.9640 & \textbf{0.9685} & \textbf{0.9612} & \textbf{0.9741} & \textbf{0.9786} & \underline{0.7831} & \textbf{0.7232} & \underline{0.8245} & \underline{0.8901} \\ \hline
\end{tabular}
}
\end{table*}

\section{Experiments}
\subsection{Experimental Setup}
\subsubsection{Datasets}
Due to the lack of THKG datasets in the literature, we extend two widely used TKG datasets (by crawling their qualifiers) and one HKG dataset (by crawling the time information), as our THKG benchmark datasets. First, \textbf{Wiki} is an extended dataset derived from the TKG dataset Wikidata11k \cite{T-GAP}, which only contains base triplets with time information of each triplet through a time-related conjunction (either \textit{OccurSince} or \textit{OccurUntil}). If two facts share the same base triplet and the same set of qualifiers but different time information with \textit{OccurSince} and \textit{OccurUntil}, we merge them into one fact with our conjunction \textit{Period}. Afterward, to complete its hyper-relational information, we query qualifiers of each triplet through Wikidata. Second, \textbf{YAGO} is an extension of the TKG dataset YAGO1830 \cite{xERTE}. This dataset has only base triplets from YAGO and it is originally presented in a temporal expansion scheme (see Figure \ref{fig:IntroFact}) with a temporal granularity of one year. We first merge the facts that share the same triplet but with consecutive timestamps (e.g., 1891, 1892, and 1893) into one fact with our conjunction \textit{Period} (with $t_1=$1891 and $t_2=$1893). We then collect the hyper-relational information of each fact by searching for its entity names on Wikipedia and record their QIDs on the Wikidata item page; for relations, we manually map them to the Wikidata item format following a predefined list \cite{ding2023exploring}; we then collect the qualifiers of each fact following the same process as for the Wiki dataset. As a few collected qualifiers here also contain time information, we update the time information of the corresponding facts according to the (latest) qualifiers we collected from Wikidata. Note that we drop the fact that cannot be mapped to Wikidata following the above process (about 0.2\%). Third, \textbf{wikipeople} is an extended dataset from the HKG dataset wikipeople \cite{NaLP}, originally formatted as an n-ary representation. We first convert its representation from n-ary to base triplets with key-value pairs, and then collect the time information for each fact if available on Wikidata; we finally filter the fact whose entities and relations have ever appeared in fact with time information. In addition, we also include one traditional TKG dataset \textbf{ICEWS14} \cite{garcia2018learning} which does not have any qualifiers, to further validate our method. As ICEWS14 originally contains political events with specific timestamps of a temporal granularity of one day, we thus merge facts to fit our versatile time representation following a similar process as for YAGO; the resulting time-related conjunctions are all \textit{Period} due to the short duration of political events. Note that it is impractical to collect any qualifiers for ICEWS14 because those events can hardly be linked to the facts of any public KGs.
% and all facts also have a exact happen time, so the facts in ICEWS14Day are always without conjunction \textit{Since}, \textit{Until}, \textit{Infty} and also qualifiers.
Table \ref{tbl:DatasetStatistic} presents the dataset statistics.

% Wikipedia\footnote{https://www.wikipedia.org/}
% interpolated format, with entities and relations presented in natural language rather than wiki identifiers. For this dataset, we retain the start and end times of each fact and discard other facts occurring within this period.

% To map entity names to wiki identifiers, we search for the names on Wikipedia\footnote{https://www.wikipedia.org/} and record their QIDs on the Wikidata item page. For relation names, we manually map them to the Wikidata item format following a predefined list \cite{ding2023exploring}; for example, \textit{graduatedFrom} is equivalent to \textit{educated at}, and \textit{isMarriedTo} is equivalent to \textit{spouse}. Finally, we collect the qualifiers of each fact following the construction of the wiki dataset. \textbf{wikipeople} is a restructured and extended dataset from the HKG dataset wikipeople \cite{xERTE}, formatted as an n-ary representation. Our first step is to convert its representation from n-ary to base triplet with key-value pairs, then collect the time information for a portion of the data that previously lacked time scope. Finally, we retain only high-frequency time-invariant facts (approximately 10\% of the training, validation, and test sets, respectively) and eliminate low-frequency facts. Table \ref{tbl:DatasetStatistic} presents the statistics of each proposed dataset. Note that we limit the number of qualifiers to five.

\subsubsection{Baselines}
We compare VITA against a sizeable collection of state-of-the-art techniques of two categories.
% , with the settings below. 
The first category includes TKG embedding models: \textbf{TeRO} \cite{Tero}, \textbf{DE-SimplE} \cite{DE-SimplE}, \textbf{BoxTE} \cite{BoxTE}, \textbf{HypeTKG} \cite{ding2023exploring}, \textbf{HGE} \cite{HGE}, \textbf{TARGCN} \cite{TARGCN}, and \textbf{literalE} \cite{literalE}. As TKG embedding models can only handle the base triplets, we thus remove the qualifiers to adapt to these methods. For methods designed for the temporal expansion representation (TeRO, DE-SimplE, BoxTE, and HypeTKG), we set the time granularity to 1 year (the finest granularity) following the existing works. For methods designed for the temporal snapshot representation (HGE, TARGCN), we also set the time interval to 1 year (the smallest time interval) to keep tracking the finest temporal dynamics. In addition, as literalE is designed for triplets with numeric literals, we adopt the reification strategy \cite{brickley2014rdf} to assign the time-related conjunction and time values to the reified base triplet; we consider literalE with three base models: \textbf{literalE-DisMult}, \textbf{literalE-ComplEx}, and \textbf{literalE-ConvE}.
The second category includes HKG embedding models: \textbf{NaLP-Fix} \cite{HINGE}, \textbf{HINGE} \cite{HINGE}, \textbf{StarE} \cite{StarE}, \textbf{GRAN} \cite{GRAN}, \textbf{HyConvE} \cite{HyConvE}, and \textbf{HypE} \cite{HyPE}, and \textbf{HyNT} \cite{HyNT}. Among these methods, only HyNT can handle numeric literals in this category, and the time information is represented as qualifiers in HyNT as shown in Figure \ref{fig:IntroFact}. For other methods in this category, we remove the time information and keep only the base triplet and qualifiers.
The details of all baselines and their hyperparameter settings are presented in Appendix \ref{appx:baselines}.

\subsubsection{Evaluation Tasks and Metrics}
We consider link prediction as our evaluation task. Following the definition of link prediction tasks for THKG in Section \ref{sec:def}, we report the results on entity prediction, relation prediction, time prediction, and numeric literal prediction. Note that some baseline methods are not designed for predicting all elements, we thus report only the results of the baselines when applicable. We consider the commonly used metrics for link prediction tasks, Mean Reciprocal Rank (\textbf{MRR}), \textbf{Hit@1}, \textbf{Hit@3}, and \textbf{Hit@10} for entity and relation prediction, and Mean Squared Error (\textbf{MSE}) for time and numeric prediction. 

% It predicts a missing element in a hyper-relational fact, such as $(s, r, ?, k_1, v_1, ...)$ or $(s, ?, o, k_1, v_1, ...)$, where the missing element is an entity or a relation, respectively. We generate a ranking list of entities or relations using each method and report four commonly used metrics, i.e., Mean Reciprocal Rank
% (\textbf{MRR}), \textbf{Hit@1}, \textbf{Hit@3}, and \textbf{Hit@10}, for link prediction tasks on entities and relations separately. Moreover, we also evaluate numeric and time predictions as regression tasks, calculating the error between the predicted and the ground-truth values. We report the Mean Squared Error (\textbf{MSE}) as the evaluation metric.

Note that for the baselines of the TKG family with temporal expansion representation, we report the average result of facts that are temporally expanded from the same fact for a fair comparison. For example, a THKG fact $\{(s, r, o), (Period, t, t+2)\}$ is temporally expanded to three facts $(s, r, o)$ with timestamps $t$, $t+1$, and $t+2$, respectively; the average performance on these three expanded facts are reported for this THKG fact.

\subsection{Comparison in Link Prediction Performance}
% We compare the performance of VITA against baselines on different prediction tasks.

\subsubsection{Comparison with Baselines in Entity Prediction.} Table \ref{tab:AllLinkPrediction} shows the results in entity prediction. The top two best-performing methods are highlighted in bold and underlined, respectively. We observe that our VITA achieves the best performance in most cases, except the Hit@10 on Wiki and YAGO where VITA is slightly lower than HyNT and HyConvE, respectively. In general compared to the best-performing baselines, VITA yields 4.4\%, 4.7\%, 8.7\%, and 5.1\% improvements on Wiki, YAGO, wikipeople, and ICEWS14, respectively. We also have some interesting findings. First, some TKG models with the temporal expansion representation face computational challenges on wikipeople due to the out-of-memory error, because of the expansion of facts with long validity. Second, compared to HyNT which represents time information as qualifiers, VITA shows an improvement of 10.0\% on average across the four datasets, which strongly supports our design of separately accommodating the time information from numeric literals, as the temporal validity of a fact is more important than other qualifiers that only provide auxiliary information about the fact.

% Overall, our model yields average improvements of 4.36\% and 9.4\% on Wiki and wikipeople datasets, respectively. Notably, compared to the best-performing baseline HyNT, consistently outperform other HKG baselines, 
% by 34.76\%, 5.06\% and 9.4\% on three datasets respectively, 
% which illustrates that learning from time information can boost the link prediction performance. Compared with HyNT, our model achieves 4.36\% improvement on average across the three datasets. This substantiates that our design ethos of separately accommodating the time representation from numeric literals effectively boosts link prediction. 
% Alongside this, it also proves the versatility of time which not only provides auxiliary information but also the validity about a fact. 
% On the other hand, compared with TKG baselines, VITA yields average improvements of 78.28\% and 9.5\% on Wiki and wikipeople datasets, respectively, only with a slight drop by xxx on the YAGO dataset, xxx.

% VITA shows a slight drop by XXX on YAGO, XXX. However, , VITA can learn auxiliary information for semantic from a universal representation that contains qualifiers while in entity prediction.

\subsubsection{Comparison with Baselines in Relation Prediction.} Table \ref{tab:AllRelationPreiction} shows the results in relation prediction. Note that many baselines are not applicable in this task because they cannot predict relations. We see that most applicable methods achieve high performance, due to the small number of relations; our VITA is among the best-performing ones.

% Table \ref{tab:AllRelationPreiction} shows the results in relation prediction. Our VITA is comparable with the best-performing baselines, with a slight decline of 0.1\% on average over the four datasets. The insignificant decline is attributed to the small number of relations, where all applicable methods achieve high performance.
 
% However, VITA is a much more precise model because it performs better by up to 1.4\% improvement on Hit@1 metrics.

\subsubsection{Comparison with Baselines in Time \& Numeric Literal Prediction.} Table \ref{tab:NumericTimePrediction} shows the results in time and numeric literal prediction (only the baseline HyNT is applicable). We observe that our VITA consistently outperforms HyNT with 75.3\% and 3.5\% improvements in predicting time values and numeric literals, respectively. In particular, VITA has a large advantage in time prediction, which benefits from our design of the versatile time representation.

% 28.51\%, 35.35\%, and 51.69\% on Wiki, YAGO, and wikipeople datasets, respectively. Particularly, VITA yields a 67.44\% improvement on average in time prediction across three datasets, elucidating that VITA has a large advantage in time representation over baselines.
% the average of numeric and time (AVG), and 4 types of time prediction. 
% We observe that our VITA consistently outperforms HyNT the AVG metric by 28.51\%, 35.35\%, and 51.69\% on Wiki, YAGO, and wikipeople datasets, respectively. Particularly, VITA yields a 67.44\% improvement on average in time prediction across three datasets, elucidating that VITA has a large advantage in time representation over baselines. 
% Additionally, we can observe that VITA shows slight improvement in numeric literals prediction but a clear improvement in time prediction, which proves our design techniques for learning time representation in a versatile time encoder achieve better performance on Temporal Hyper-relational KGs.

\begin{table}
\caption{Link prediction performance on time and numeric literals (in MSE). Note that ``N/A'' appears because the ICEWS14 dataset does not contain any numeric literal.}
\vspace{-1em}
\label{tab:NumericTimePrediction}
% \small
\resizebox{\linewidth}{!}{
\begin{tabular}{l|cc|cc|cc|cc}
\hline
\multicolumn{1}{c|}{\multirow{2}{*}{Method}} & \multicolumn{2}{c|}{Wiki} & \multicolumn{2}{c|}{YAGO} & \multicolumn{2}{c|}{wikipeople} & \multicolumn{2}{c}{ICEWS14} \\ \cline{2-9} 
& Num & Time & Num & Time & Num & Time & Num & Time \\ \hline
HyNT & 0.0537 & 0.0355 & 0.0657 & 0.0339 & 0.0447 & 0.0487 & N/A & 0.2871 \\
VITA & \textbf{0.0508} & \textbf{0.0108} & \textbf{0.0624} & \textbf{0.0117} & \textbf{0.0446} & \textbf{0.0160} & N/A & \textbf{0.0033} \\
\hline
\end{tabular}
}
\end{table}

% \begin{table}[]
% \caption{Link prediction performance on time and numeric literals.}
% \label{tab:NumericTimePrediction}
% % \resizebox{\columnwidth}{!}{%
% % \footnotesize
% \small 
% \begin{tabular}{l|l|ccccccc}
% \hline
% \multicolumn{1}{l|}{Dataset} & Method & AVG & Num & Time & Type1 & Type2 & Type3 & Type4 \\ \hline
% \multirow{2}{*}{Wiki} & HyNT & 0.0450 & 0.0537 & 0.0355 & 0.0379 & 0.0335 & 0.0312 & 0.0155 \\
%  & VITA & 0.0321 & 0.0508 & 0.0108 & 0.0115 & 0.0081 & 0.0293 & 0.0159 \\ \hline
% \multirow{2}{*}{YAGO} & HyNT & 0.0425 & 0.0657 & 0.0339 & 0.0345 & 0.0334 & 0.0354 & 0.0193 \\
%  & VITA & 0.0275 & 0.0624 & 0.0117 & 0.0104 & 0.0100 & 0.0349 & 0.0267 \\ \hline
% \multirow{2}{*}{wikipeople} & HyNT & 0.0483 & 0.0447 & 0.0487 & 0.0451 & 0.0433 & 0.0921 & 0.0222 \\
%  & VITA & 0.0233 & 0.0446 & 0.0160 & 0.0068 & 0.0069 & 0.0552 & 0.0202 \\ \hline
% \end{tabular}%
% % }
% \end{table}

\subsection{Ablation Study}
To further validate the design choices of our VITA, we conduct ablation studies to evaluate the effectiveness of the versatile time representation and a key component of VITA, TimeSpan Fuser.

\begin{table*}
\caption{Ablation study on entity prediction.}
\vspace{-1em}
\label{tab:EntAblation}
% \small
\resizebox{\linewidth}{!}{
\begin{tabular}{l|cccc|cccc|cccc|cccc}
\hline
\multicolumn{1}{c|}{\multirow{2}{*}{Method}} & \multicolumn{4}{c|}{Wiki} & \multicolumn{4}{c|}{YAGO} & \multicolumn{4}{c|}{wikipeople} & \multicolumn{4}{c}{ICEWS14} \\ \cline{2-17} 
\multicolumn{1}{c|}{} & MRR & Hit@1 & Hit@3 & Hit@10 & MRR & Hit@1 & Hit@3 & Hit@10 & MRR & Hit@1 & Hit@3 & Hit@10 & MRR & Hit@1 & Hit@3 & Hit@10 \\ \hline
\textit{w/o VTR} & 0.3638 & 0.2895 & 0.3838 & \textbf{0.5194} & 0.1791 & 0.1188 & 0.1984 & \textbf{0.3002} & 0.2527 & 0.1836 & 0.2706 & 0.3922 & 0.5392 & 0.4321 & 0.6071 & \textbf{0.7379} \\
\textit{w/o TSF} & 0.3655 & 0.2939 & 0.3871 & 0.5074 & 0.1830 & 0.1364 & 0.1934 & 0.2777 & 0.2629 & 0.1993 & 0.2800 & \textbf{0.3934} & 0.5383 & 0.4313 & 0.6088 & 0.7339 \\
VITA & \textbf{0.3780} & \textbf{0.3113} & \textbf{0.3926} & 0.5168 & \textbf{0.1898} & \textbf{0.1366} & \textbf{0.2049} & 0.2993 & \textbf{0.2631} & \textbf{0.2012} & \textbf{0.2801} & 0.3897 & \textbf{0.5450} & \textbf{0.4396} & \textbf{0.6132} & 0.7361 \\ \hline
\end{tabular}%
}
\end{table*}

\begin{table*}
\caption{Ablation study on relation prediction.}
\vspace{-1em}
\label{tab:RelAblation}
% \small
\resizebox{\linewidth}{!}{%
\begin{tabular}{l|cccc|cccc|cccc|cccc}
\hline
\multicolumn{1}{c|}{\multirow{2}{*}{Method}} & \multicolumn{4}{c|}{Wiki} & \multicolumn{4}{c|}{YAGO} & \multicolumn{4}{c|}{wikipeople} & \multicolumn{4}{c}{ICEWS14} \\ \cline{2-17} 
\multicolumn{1}{c|}{} & MRR & Hit@1 & Hit@3 & Hit@10 & MRR & Hit@1 & Hit@3 & Hit@10 & MRR & Hit@1 & Hit@3 & Hit@10 & MRR & Hit@1 & Hit@3 & Hit@10 \\ \hline
\textit{w/o VTR} & 0.9778 & 0.9731 & 0.9817 & 0.9842 & 0.9202 & 0.8922 & 0.9418 & 0.9578 & 0.9332 & 0.9109 & 0.9503 & 0.9652 & 0.7823 & 0.7203 & \textbf{0.8266} & \textbf{0.8930} \\
\textit{w/o TSF} & 0.9808 & 0.9761 & 0.9843 & 0.9870 & 0.9220 & 0.8939 & 0.9471 & 0.9637 & 0.9490 & 0.9371 & 0.9578 & 0.9661 & 0.7712 & 0.7103 & 0.8119 & 0.8843 \\
VITA & \textbf{0.9856} & \textbf{0.9823} & \textbf{0.9880} & \textbf{0.9897} & \textbf{0.9462} & \textbf{0.9346} & \textbf{0.9531} & \textbf{0.9640} & \textbf{0.9685} & \textbf{0.9612} & \textbf{0.9741} & \textbf{0.9786} & \textbf{0.7831} & \textbf{0.7232} & 0.8245 & 0.8901 \\ \hline
\end{tabular}%
}
\end{table*}

% VITA(noMD \& VTE) & \multicolumn{1}{c|}{0.3660} & 0.9778 & \multicolumn{1}{c|}{0.1883} & 0.9142 & \multicolumn{1}{c|}{0.2462} & 0.9236 \\
% VITA(noMD) & \multicolumn{1}{c|}{0.3759} & 0.9887 & \multicolumn{1}{c|}{0.1846} & 0.9526 & \multicolumn{1}{c|}{0.2636} & 0.9690 \\
% VITA(abl1) & \multicolumn{1}{c|}{0.3705} & 0.9856 & \multicolumn{1}{c|}{0.1853} & 0.9447 & \multicolumn{1}{c|}{0.2610} & 0.9185 \\
% VITA(abl2) & \multicolumn{1}{c|}{0.3743} & 0.9852 & \multicolumn{1}{c|}{0.1889} & 0.9469 & \multicolumn{1}{c|}{0.2652} & 0.9681 \\

\begin{table}
\caption{Ablation study on time and numeric literal prediction (in MSE).}
\vspace{-1em}
\label{tab:NumTimeAblation}
% \small
\resizebox{\linewidth}{!}{
\begin{tabular}{l|cc|cc|cc|cc}
\hline
\multirow{2}{*}{Method} & \multicolumn{2}{c|}{Wiki} & \multicolumn{2}{c|}{YAGO} & \multicolumn{2}{c|}{wikipeople} & \multicolumn{2}{c}{ICEWS14} \\ \cline{2-9} 
 & \multicolumn{1}{c}{Num} & Time & \multicolumn{1}{c}{Num} & Time & \multicolumn{1}{c}{Num} & Time & \multicolumn{1}{c}{Num} & Time \\ \hline
\textit{w/o VTR} & 0.0528 & 0.0159 & \textbf{0.0536} & 0.0339 & 0.0486 & 0.0347 & N/A	& 0.0204 \\
\textit{w/o TSF} & \multicolumn{1}{c}{0.0513} & 0.0224 & \multicolumn{1}{c}{0.0635} & 0.0134 & \multicolumn{1}{c}{0.0448} & 0.0265 & N/A & 0.0058 \\
VITA & \textbf{0.0508} & \textbf{0.0108} & \multicolumn{1}{c}{0.0624} & \textbf{0.0117} & \textbf{0.0446} & \textbf{0.0160} & N/A	& 0.0033 \\ \hline
\end{tabular}%
}
\vspace{-1em}
\end{table}

% VITA(noMD \& VTE) & \multicolumn{1}{c|}{0.0544} & 0.0151 & \multicolumn{1}{c|}{0.0528} & 0.0200 & \multicolumn{1}{c|}{0.0437} & 0.0255 \\
% VITA(noMD) & \multicolumn{1}{c|}{0.0512} & 0.0090 & \multicolumn{1}{c|}{0.0626} & 0.0132 & \multicolumn{1}{c|}{0.0465} & 0.0150 \\
% VITA(abl1) & \multicolumn{1}{c|}{0.0553} & 0.3242 & \multicolumn{1}{c|}{0.0736} & 0.0387 & \multicolumn{1}{c|}{0.0461} & 0.0392 \\
% VITA(abl2) & \multicolumn{1}{c|}{0.0535} & 0.0932 & \multicolumn{1}{c|}{0.0518} & 0.0364 & \multicolumn{1}{c|}{0.0441} & 0.3306 \\

\subsubsection{Impact of the versatile time representation.}    
We verify the utility of our versatile time representation by designing a variant \textit{w/o VTR}, which removes the specific pipeline for learning from time triplets, but instead regards the time information as numeric literals. We then use the qualifier encoder to encode time values and employ the qualifier prediction Transformer as the decoder for link predictions. Tables \ref{tab:EntAblation}, \ref{tab:RelAblation} and \ref{tab:NumTimeAblation} present the prediction results. We observe that the versatile time representation scheme effectively improves the link prediction performance of VITA, in particular, achieving 3.6\% and 58.8\% improvements on entity and time prediction, respectively.

\subsubsection{Impact of the TimeSpan Fuser.} 
We study the effectiveness of TimeSpan Fuser considering a variant \textit{w/o TSF}, where the TSF component is removed from the decoder and the context feature from the encoder is directly fed to the corresponding decoder for prediction. The results are shown in Table \ref{tab:EntAblation}, \ref{tab:RelAblation} and \ref{tab:NumTimeAblation}. We observe that the TSF boosts the performance of VITA on all link prediction tasks. In particular, TSF brings average improvements of 2.2\% and 36.8\% in predicting entities and time, respectively. This indicates that further integrating the validity timespan information into context features is vital for enhancing link prediction performance.

% and \textit{w/o VTR} predicts the same entity with rank 6; 

\subsection{Case Study}
\begin{figure}
    \centering
    \includegraphics[width=\linewidth]{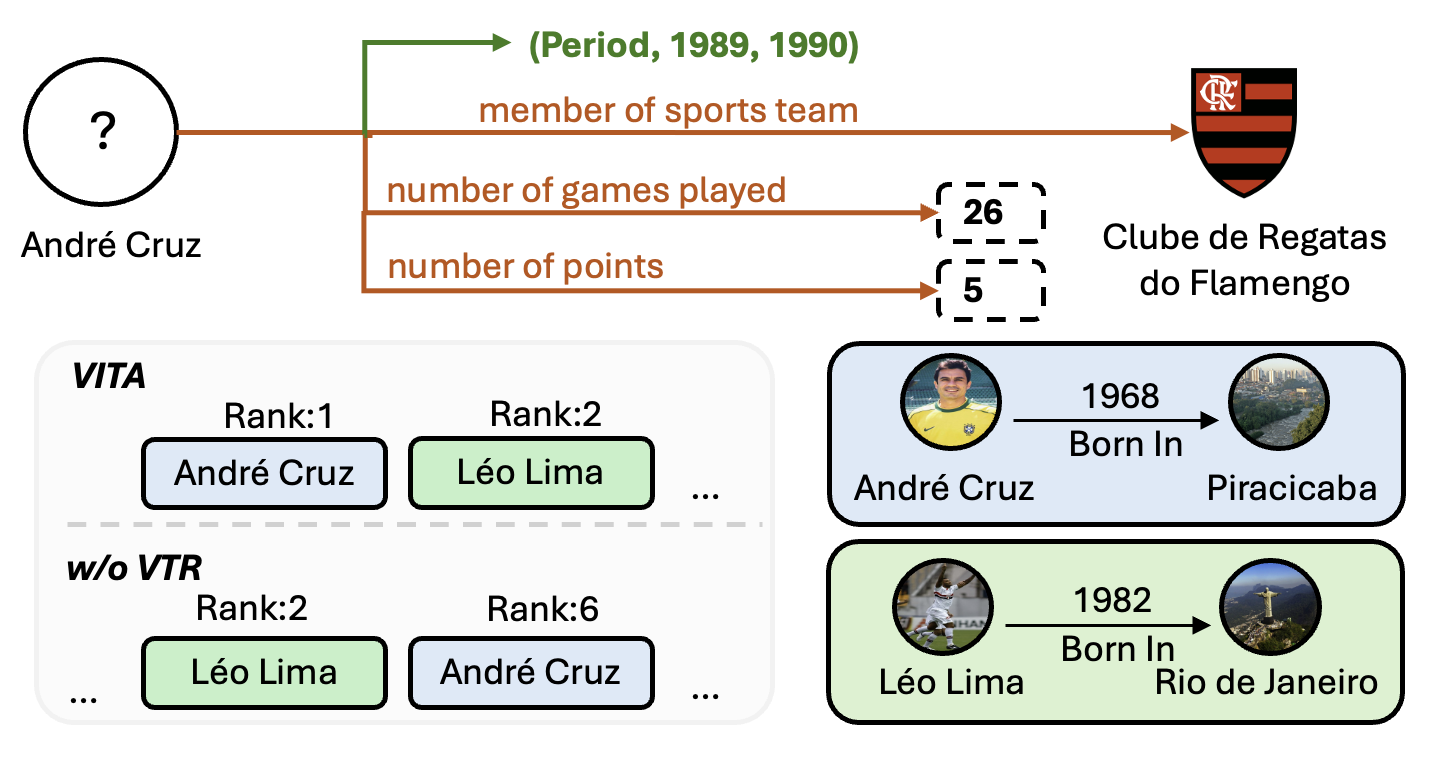}
    \vspace{-2em}
    \Description[<short description>]{<long description>}
    \caption{Case Study.}
    \label{fig:CaseStudy}
    \vspace{-1.5em}
\end{figure}

We conduct a case study to demonstrate the benefit of VITA in boosting the link prediction performance. As shown in Figure \ref{fig:CaseStudy}, a test query fact on YAGO is \textit{\{(?, member of sports team, Clube de Regatas do Flamengo, (Period, 1989, 1990), \{(number of games played, 26), (number of points, 5))\} \}}, where the missing element is the subject and the ground truth is ``André Cruz''. The prediction results of VITA and its ablated variant \textit{w/o VTR} show that VITA predicts the correct answer ``André Cruz'' with rank 1 and another similar entity ``Léo Lima'' with rank 2; in contrast, \textit{w/o VTR} predicts ``Léo Lima'' with rank 2 and the ground truth ``André Cruz'' with rank 6. In the training dataset, we find two facts related to the birth information about the two entities, \textit{\{(André Cruz, born in, Piracicaba), (period, 1968, 1968) \}} and \textit{\{(Léo Lima, born in, Rio de Janeiro), (period, 1982, 1982) \}}, respectively; the birth year serves as a strong clue to rank André Cruz in front of Léo Lima, because of the reasonable age of players being a member of a professional sport team. VITA (with versatile time representation) captures this subtle clue and ranks André Cruz in front of Léo Lima, while \textit{w/o VTR} (representing time information as qualifiers) fails in this case.

% the improvement of applying Time Value Encoder (TVE) and TimeSpan Fuser (TSF). Consider a link prediction query of THKG, \{(?, member of sports team, Clube de Regatas do Flamengo, (Period, 1989, 1990), (number of matches played/races/starts, 26), (number of points/goals/set scored, 5))\}, we aim to predict the missing subject. 

% The results of HyNT and VITA show that VITA predict the correct answer “André Cruz” with rank 1 and HyNT only predict it with rank 6. We notice that some answers in HyNT with higher rank seem unreasonable, for example, we have learned that the second rank answer “Léo Lima” born in 1982, it should have a low probability compared with “André Cruz” because we learned that “André Cruz” born in 1968, that is “André Cruz” with 21 years old is more rational than “Léo Lima” with only 7 years old. Therefore, the validity of fact can be considered more in VITA.

\section{Conclusion}
In this paper, we introduce VITA, a \underline{V}ersatile t\underline{I}me represen\underline{TA}tion learning method for Temporal Hyper-relational Knowledge Graphs. We first revisit the limitation of the existing time representation for temporal hyper-relational knowledge graphs, and propose a versatile time representation that can flexibly accommodate different types of temporal validity of facts (i.e., since, until, period, time-invariant), benefiting from the advantages of preciseness, compactness, and completeness of the time representation. Subsequently, VITA is designed to effectively learn from the temporal hyper-relational facts under our versatile time representation to boost the link prediction performance. Following an encoder-decoder architecture, VITA learns the time information not only from time values of both real-valued numbers or infinity tokens, but also from the timespan information prescribing the validity timespan of facts. We conduct a thorough evaluation of VITA compared to a sizable collection of baselines on real-world KG datasets. Results show that VITA outperforms the best-performing baselines in various link prediction tasks (predicting missing entities, relations, time, and numeric literals) by up to 75.3\%. Ablation studies and a case study also verify our key design choices.
    
In the future, we plan to investigate multi-modal temporal HKGs, such as images connecting to entities evolving over time.


% \begin{acks}
% \end{acks}

\bibliographystyle{ACM-Reference-Format}
\bibliography{sample-base}

\appendix

\section{Details of Baselines and Settings}
\label{appx:baselines}
We compare VITA against a sizeable collection of state-of-the-art techniques of two categories. 

\noindent The first category includes TKG embedding models: 

\begin{itemize}[leftmargin=*]

\item \textbf{TeRO}\footnote{\url{https://github.com/soledad921/ATISE}} \cite{Tero} is a time-aware knowledge graph embedding model that incorporates temporal information by applying rotational transformations in complex space. We empirically set the embedding dimension, learning rate, and minimum threshold to \{500, 0.1, 100\}.

\item \textbf{DE-SimplE}\footnote{\url{https://github.com/BorealisAI/de-simple}} \cite{DE-SimplE} employs diachronic embeddings to capture the evolving semantics of entities and relations in temporal knowledge graphs. We empirically set the batch size, learning rate, embedding size, dropout, and training epoch to \{512, 0.001, 100, 0.4, 500\}.

\item \textbf{BoxTE}\footnote{\url{https://github.com/JohannesMessner/BoxTE}} \cite{BoxTE} leverages box embeddings to represent temporal knowledge graphs, capturing temporal uncertainty and hierarchical structures within a geometric space. We empirically set the margin, training epoch, batch size, embedding size, and learning rate to \{0.2, 10, 128, 300, 0.0001\}.

\item \textbf{HypeTKG}\footnote{\url{https://github.com/0sidewalkenforcer0/HypeTKG}} \cite{ding2023exploring} enhances link prediction in hyper-relational temporal knowledge graphs by integrating time-invariant relational knowledge. We empirically set the batch size, embedding size, training epoch, and learning rate to \{256, 300, 400, 0.0001\}.

\item \textbf{HGE}\footnote{\url{https://github.com/NacyNiko/HGE}} \cite{HGE} embeds temporal knowledge graphs into a product space composed of heterogeneous geometric subspaces, capturing diverse relational patterns and temporal dynamics. We empirically set the training epoch, batch size, and learning rate to \{200, 1000, 0.1\}.

\item \textbf{TARGCN}\footnote{\url{https://github.com/ZifengDing/TARGCN}} \cite{TARGCN} introduces a straightforward yet effective graph encoder for temporal knowledge graph completion, capturing both structural and temporal dependencies. We empirically set the embedding size, the number of aggregation steps, the activation function, the search range, and the number of temporal neighbors to \{300, 1, Tanh, 365, 100\}.

\item \textbf{literalE}\footnote{\url{https://github.com/SmartDataAnalytics/LiteralE}} \cite{literalE} enhances knowledge graph embeddings by incorporating literal information, such as numerical and textual attributes, to enrich entity representations. We empirically set the embedding size, batch size, training epoch, and learning rate to \{100, 128, 100, 0.001\} for LiteralE-DisMult and LiteralE-ComplEx, and to \{200, 128, 100, 0.001\} for LiteralE-ConvE.
\end{itemize}

\noindent The second category includes HKG embedding models:

\begin{itemize}[leftmargin=*]

\item \textbf{NaLP-Fix}\footnote{\url{https://github.com/eXascaleInfolab/HINGE_code/tree/master/NALP}} \cite{NaLP} captures the interactions between relation-entity pairs using CNNs. We empirically set the batch size, embedding size, training epoch, and learning rate to \{128, 100, 400, 0.00005\}.

\item \textbf{HINGE}\footnote{\url{https://github.com/eXascaleInfolab/HINGE_code}} \cite{HINGE} repeatedly learns from triplets and affiliated key-value pairs using CNNs. We empirically set the batch size, embedding size, the number of convolution filers, and learning rate to \{128, 100, 400, 0.0001\}.

\item \textbf{StarE}\footnote{\url{https://github.com/migalkin/StarE}} \cite{StarE} designs a directed heterogeneous graph encoder to capture the interactions among elements in a fact. We empirically set the batch size, embedding size, graph layers, learning rate, and training epoch to \{128, 200, 2, 0.0001, 400\}.

\item \textbf{GRAN}\footnote{\url{https://github.com/PaddlePaddle/Research/tree/master/KG/ACL2021_GRAN}} \cite{GRAN} incorporates edge biases to discriminate connections between elements in a fact and harnesses the self-attention mechanism to further capture the correlation. We empirically set the batch size, embedding size, training epoch, and learning rate to \{1024, 256, 100, 0.0005\}.

\item \textbf{HyConvE}\footnote{\url{https://github.com/CarllllWang/HyConvE}} \cite{HyConvE} leverages 3D convolution to capture the sophisticated interactions among entities and relations in a fact. We empirically set the learning rate, batch size, embedding size, and training epoch to \{0.01, 128, 400, 500\}.

\item \textbf{HypE}\footnote{\url{https://github.com/ServiceNow/HypE}} \cite{HyPE} learns a unified representation for each entity and relation, leveraging positional convolutional weight filters for every position in a fact. We empirically set the training epoch, batch size, learning rate, and embedding size to \{1000, 128, 0.1, 200\}.

\item \textbf{HyNT}\footnote{\url{https://github.com/bdi-lab/HyNT}} \cite{HyNT} develops a context Transformer to learn representations of the primary triplets and the qualifiers by exchanging information among them. We empirically set the training epoch, batch size, learning rate, embedding size, and dropout to \{1050, 1024, 0.0004, 256, 0.15\}.

\end{itemize}

\end{document}